\documentclass{article}

\usepackage[square,numbers]{natbib}
\usepackage[final]{./nips_2017_nofooter}

\usepackage[utf8]{inputenc} 
\usepackage[T1]{fontenc}    
\usepackage{hyperref}       
\usepackage{url}            
\usepackage{booktabs}       
\usepackage{amsfonts}       
\usepackage{nicefrac}       
\usepackage{microtype}      
\usepackage{graphicx}
\usepackage{color}
\usepackage{amsmath}
\usepackage{subcaption}
\usepackage{algorithm}
\usepackage{algorithmic}
\usepackage{color}

\usepackage{sidecap}
\usepackage{subcaption}
\usepackage{wrapfig}

\usepackage{color}

\newcommand{\walls}{\emph{slalom walls}}
\newcommand{\heightfields}{\emph{variable terrain}}
\newcommand{\hurdles}{\emph{hurdles}}
\newcommand{\obstacles}{\emph{obstacles}}
\newcommand{\gaps}{\emph{gaps}}
\newcommand{\platforms}{\emph{platforms}}
\newcommand{\walker}{\emph{Planar Walker}}
\newcommand{\quadruped}{\emph{Quadruped}}
\newcommand{\humanoid}{\emph{Humanoid}}
\newcommand{\reacher}{\emph{Memory Reacher}}

\def\EE{\mathbb{E}}

\title{Emergence of Locomotion Behaviours \\in Rich Environments}

\author{
  Nicolas Heess, Dhruva TB, Srinivasan Sriram, Jay Lemmon, Josh Merel, Greg Wayne, \\
  \textbf{Yuval Tassa, Tom Erez, Ziyu Wang, S. M. Ali Eslami, Martin Riedmiller,  David Silver} \\
  DeepMind\\
}

\def\EE{\mathbb{E}}

\begin{document}
\newcommand{\razp}[1]{\textbf{\color{magenta}{[Razp: #1]}}}

\maketitle
 
\begin{abstract}
The reinforcement learning paradigm allows, in principle, for complex behaviours to be learned directly from simple reward signals. In practice, however, 
it is common to carefully hand-design the reward function to encourage a particular solution, or to derive it from demonstration data. In this paper explore how a rich environment can help to promote the learning of complex behavior.
Specifically, we train agents in diverse environmental contexts, and find that this encourages the emergence of robust behaviours that perform well across a suite of tasks. We demonstrate this principle for locomotion -- behaviours that are known for their sensitivity to the choice of reward. We train several simulated bodies on a diverse set of challenging terrains and obstacles, using a simple reward function based on forward progress. Using a novel scalable variant of policy gradient reinforcement learning, our agents learn to run, jump, crouch and turn as required by the environment without explicit reward-based guidance. A visual depiction of highlights of the learned behavior can be viewed in \href{https://youtu.be/hx_bgoTF7bs}{{\color{blue}this video}}.
\end{abstract}
 
\section{Introduction} 
 
Reinforcement learning has demonstrated remarkable progress, achieving high levels of performance in Atari games \cite{Mnih2015Human}, 3D navigation tasks \cite{MniBadMir2016a,jaderberg2016reinforcement}, and board games \cite{silver2016mastering}. What is common among these tasks is that there is a well-defined reward function, such as the game score, which can be optimised to produce the desired behaviour. However, there are many other tasks where the ``right'' reward function is less clear, and optimisation of a na\"{i}vely selected one can lead to surprising results that do not match the expectations of the designer. This is particularly prevalent in continuous control tasks, such as locomotion, and it has become standard practice to carefully handcraft the reward function, or else elicit a reward function from demonstrations.

Reward engineering has led to a number of successful demonstrations of locomotion behaviour, however, these examples are known to be brittle: they can lead to unexpected results if the reward function is modified even slightly, and for more advanced behaviours the appropriate reward function is often non-obvious in the first place. Also, arguably, the requirement of careful reward design sidesteps a primary challenge of reinforcement learning: how an agent can learn for itself, directly from a limited reward signal, to achieve rich and effective behaviours. In this paper we return to this challenge. 
 
Our premise is that rich and robust behaviours will emerge from simple reward functions, if the environment itself contains sufficient richness and diversity. Firstly, an environment that presents a spectrum of challenges at different levels of difficulty may shape learning and guide it towards solutions that would be difficult to discover in more limited settings. Secondly, the sensitivity to reward functions and other experiment details may be due to a kind of overfitting, finding idiosyncratic solutions that happen to work within a specific setting, but are not robust when the agent is exposed to a wider range of settings. Presenting the agent with a diversity of challenges thus increases the performance gap between different solutions and may favor the learning of solutions that are robust across settings.

We focus on a set of novel locomotion tasks that go significantly beyond the previous state-of-the-art for agents trained directly from reinforcement learning. They include a variety of obstacle courses for agents with different bodies (\quadruped{}, \walker{}, and \humanoid{} \cite{tassa2012synthesis,schulman2015high}).
The courses are procedurally generated such that every episode presents a different instance of the task.

Our environments include a wide range of obstacles with varying levels of difficulty (e.g. steepness, unevenness, distance between gaps). The variations in difficulty present an implicit curriculum to the agent -- as it increases its capabilities it is able to overcome increasingly hard challenges, resulting in the emergence of ostensibly sophisticated locomotion skills which may na\"ively have seemed to require careful reward design or other instruction. We also show that learning speed can be improved by explicitly structuring terrains to gradually increase in difficulty so that the agent faces easier obstacles first and harder obstacles only when it has mastered the easy ones.

In order to learn effectively in these rich and challenging domains, it is necessary to have a reliable and scalable reinforcement learning algorithm. We leverage components from several recent approaches to deep reinforcement learning. First, we build upon robust policy gradient algorithms, such as trust region policy optimization (TRPO) and proximal policy optimization (PPO) \cite{Schulman2015Trust,PPO_talk}, which bound parameter updates to a trust region to ensure stability. Second, like the widely used A3C algorithm \cite{MniBadMir2016a} and related approaches \cite{jaderberg2016reinforcement} we distribute the computation over many parallel instances of agent and environment. 
Our distributed implementation of PPO improves over TRPO  in terms of wall clock time with little difference in robustness, and also improves over our existing implementation of A3C with continuous actions when the same number of workers is used.

The paper proceeds as follows. In Section \ref{sec:distPPO} we describe the distributed PPO (DPPO) algorithm that enables the subsequent experiments, and validate its effectiveness empirically. Then in Section \ref{sec:Methods} we introduce the main experimental setup: a diverse set of challenging terrains and obstacles.  We provide 
evidence in Section \ref{sec:Results} that effective locomotion behaviours emerge directly from simple rewards; furthermore we show that terrains with a ``curriculum” of difficulty encourage much more rapid progress, and that agents trained in more diverse conditions can be more robust.

\section{Large scale reinforcement learning with Distributed PPO}
\label{sec:distPPO}

Our focus is on reinforcement learning in rich simulated environments with continuous state and action spaces. We require algorithms that are robust across a wide range of task variation, and that scale effectively to challenging domains. We address each of these issues in turn.

\paragraph{Robust policy gradients with Proximal Policy Optimization}

Deep reinforcement learning algorithms based on large-scale, high-throughput optimization methods, have produced state-of-the-art results in discrete and low-dimensional action spaces, e.g.\ on Atari games \cite{nair2015massively} and 3D navigation \cite{MniBadMir2016a,jaderberg2016reinforcement}. In contrast, many prior works on continuous action spaces (e.g.\ \cite{levine2014learning,Schulman2015Trust,levine2015end,lillicrap2015continuous,schulman2015high,heess2015learning}), although impressive, have focused on comparatively small problems, and the use of large-scale, distributed optimization is less widespread and the corresponding algorithms are less well developed (but see e.g.\ \cite{Wang2016Sample,gu2016deep,popov2017data}). We present a robust policy gradient algorithm, suitable for high-dimensional continuous control problems, that can be scaled to much larger domains using distributed computation.

Policy gradient algorithms \cite{williams1992simple} provide an attractive paradigm for continuous control.
They operate by directly maximizing the expected sum of rewards $J(\theta) = \EE_{\rho_\theta(\tau)}\left[ \sum_t \gamma^{t-1} r(s_t,a_t) \right]$ with respect to the parameters $\theta$ of the stochastic policy $\pi_\theta(a|s)$. The expectation is with respect to the distribution of trajectories $\tau=(s_0,a_0,s_1,\dots)$ induced jointly by the policy $\pi_\theta$ and the system dynamics $p(s_{t+1} | s_t, a_t)$: $\rho_\theta(\tau) = p(s_0)\pi(a_0|s_0)p(s_1|s_0,a_0)\dots$. The gradient of the objective with respect to $\theta$ is given by 
$\nabla_\theta J = \EE_{\theta}\left[ \sum_t \nabla_\theta \log \pi_\theta(a_t|s_t) (R_t - b_t) \right]$, where $R_t = \sum_{t'=t} \gamma^{t'-t} r(s_{t'},a_{t'})$ and $b_t$ is an  baseline that does not depend on $a_t$ or future states and actions. The baseline 
is often chosen to be $b_t= V^\theta(s_t) = \EE_\theta\left[ R_t | s_t \right]$. In practice the expected future return is typically approximated with a sample rollout and $V^{\theta}$ is replaced by a learned approximation $V_\phi(s)$ with parameters $\phi$.

Policy gradient estimates can have high variance (e.g.\ \cite{Duan2016Benchmarking}) and algorithms can be sensitive to the settings of their hyperparameters. 
Several approaches have been proposed to make policy gradient algorithms more robust. One effective measure is to employ a \emph{trust region} constraint that restricts the amount by which any update is allowed to change the policy \cite{peters2010relative,Schulman2015Trust,Wang2016Sample}. A popular algorithm that makes use of this idea is trust region policy optimization (TRPO; \cite{Schulman2015Trust}). In every iteration given current parameters $\theta_{\mathrm{old}}$, TRPO collects a (relatively large) batch of data and optimizes the surrogate loss 
%
$J_{TRPO}(\theta) = \EE_{\rho_{\theta_\mathrm{old}}(\tau)}\left[ \sum_t \gamma^{t-1} \frac{\pi_\theta(a_t | s_t)}{\pi_{\theta_\mathrm{old}}(a_t|s_t)} A^{\theta_{\mathrm{old}}}(a_t,s_t) \right]$
subject to a constraint on how much the policy is allowed to change, expressed in terms of the Kullback-Leibler divergence (KL) $\mathrm{KL}\left[\pi_{\theta_\mathrm{old}} | \pi_\theta\right] < \delta$. 
$A^\theta$ is the advantage function given as $A^\theta(s_t,a_t) = \EE_\theta\left[ R_t | s_t,a_t \right] - V^{\theta}(s_t)$. 

The Proximal Policy Optimization (PPO) algorithm \cite{PPO_talk}
can be seen as an approximate version of TRPO that relies only on first order gradients, making it more convenient to use with recurrent neural networks (RNNs) and in a large-scale distributed setting. The trust region constraint is implemented via a regularization term. The coefficient of this regularization term is adapted depending on whether the constraint had previously been violated or not (a similar idea but without the adaptive coefficient has also been used \cite{heess2015learning}). Algorithm Box \ref{alg:PPO} shows the core PPO algorithm in pseudo-code. 

\begin{algorithm}
\caption{Proximal Policy Optimization (adapted from \cite{PPO_talk})}
\label{alg:PPO}
\footnotesize
\begin{algorithmic}
    \FOR {$i \in \{1, \cdots, N\}$}
        \STATE Run policy $\pi_{\theta}$ for $T$ timesteps, collecting $\{s_t,a_t,r_t\}$
        \STATE Estimate advantages $\hat{A}_t = \sum_{t' > t} \gamma^{t'-t} r_{t'} - V_\phi(s_t)$
        \STATE $\pi_\mathrm{old} \leftarrow \pi_\theta$
        \FOR {$j \in \{1, \cdots, M\}$}
            \STATE $J_{PPO}(\theta) =  \sum_{t=1}^T \frac{\pi_{\theta}(a_t|s_t)}{\pi_{old}(a_t|s_t)} \hat{A}_t - \lambda \mathrm{KL}[\pi_{old}|\pi_{\theta}] $
            \STATE Update $\theta$ by a gradient method w.r.t. $J_{PPO}(\theta)$ 
		\ENDFOR
		\FOR {$j \in \{1, \cdots, B\}$}
            
            \STATE $L_{BL}(\phi) =  -\sum_{t=1}^T (\sum_{t' > t} \gamma^{t'-t} r_{t'} - V_\phi(s_t))^2$ 
            \STATE Update $\phi$ by a gradient method w.r.t. $L_{BL}(\phi)$ 
		\ENDFOR
		\IF {$\mathrm{KL}[\pi_{old}|\pi_{\theta}] > \beta_{high}\mathrm{KL}_{target}$}
		\STATE  $\lambda \leftarrow \alpha \lambda$
		\ELSIF {$\mathrm{KL}[\pi_{old}|\pi_{\theta}] < \beta_{low}\mathrm{KL}_{target}$}
		\STATE  $\lambda \leftarrow  \lambda/\alpha $
		\ENDIF
	\ENDFOR	
\end{algorithmic}
\end{algorithm}

In algorithm \ref{alg:PPO}, the hyperparameter $\mathrm{KL}_{target}$ is the desired change in the policy per iteration. The scaling term $\alpha>1$ controls the adjustment of the KL-regularization coefficient if the actual change in the policy stayed significantly below or significantly exceeded the target KL (i.e.\ falls outside the interval $[\beta_{low}\mathrm{KL}_{target}, \beta_{high}\mathrm{KL}_{target}]$). 

\paragraph{Scalable reinforcement learning with Distributed PPO}
To achieve good performance in rich, simulated environments, we have implemented a distributed version of the PPO algorithm (DPPO). 
Data collection and gradient calculation are distributed over workers. We have experimented with both synchronous and asynchronous updates and have found that averaging gradients and applying them synchronously leads to better results in practice.

The original PPO algorithm estimates advantages using the complete sum of rewards. To facilitate the use of RNNs with batch updates while also supporting variable length episodes we follow a strategy similar to \cite{MniBadMir2016a} and use truncated backpropagation through time with a window of length $K$. This makes it natural (albeit not a requirement) to use $K$-step returns also for estimating the advantage, i.e.\ we sum the rewards over the same $K$-step windows and bootstrap from the value function after $K$-steps: $\hat{A}_t = \sum_{i=1}^K \gamma^{i-1} r_{t+i} ~~ + ~~ \gamma^{K-1} V_\phi(s_{t+K}) - V_\phi(s_t)$.

The publicly available implementation of PPO by John Schulman
\cite{schulmanPPOcode}
adds several modifications to the core algorithm. These include normalization of inputs and rewards as well as an additional term in the loss that penalizes large violations of the trust region constraint. We adopt similar augmentations in the distributed setting but find that sharing and synchronization of various statistics across workers requires some care. 
The implementation of our distributed PPO (DPPO) is in TensorFlow, the parameters reside on a parameter server, and workers synchronize their parameters after every gradient step. Pseudocode and further details are provided in the supplemental material.

\subsection{Evaluation of Distributed PPO}
\label{sec:EvalDPPO}

We compare DPPO to several baseline algorithms. The goal of these experiments is primarily to establish that the algorithm allows robust policy optimization with limited parameter tuning and that the algorithm scales effectively. We therefore perform the comparison on a selected number of benchmark tasks related to our research interests, and compare to two algorithmic alternatives: TRPO and continuous A3C. For details of the comparison please see the supplemental material.

\paragraph{Benchmark tasks}
We consider three continuous control tasks for benchmarking the algorithms.  All environments rely on the Mujoco physics engine \cite{todorov2012mujoco}. Two tasks are locomotion tasks in obstacle-free environments and the third task is a planar target-reaching task that requires memory.  \emph{Planar walker}: a simple bipedal walker with 9 degrees-of-freedom (DoF) and 6 torque actuated joints. It receives a primary reward proportional to its forward velocity, additional terms penalize control and the violation of box constraints on torso height and angle. Episodes are terminated early when the walker falls. \emph{Humanoid}: The humanoid has 28 DoF and 21 acutated joints. The humanoid, too, receives a reward primarily proportional to its velocity along the x-axis, as well as a constant reward at every step that, together with episode termination upon falling, encourage it to not fall. \emph{Memory reacher}: A random-target reaching task with a simple 2 DoF robotic arm confined to the plane. The target position is provided for the first 10 steps of each episode during which the arm is not allowed to move.  When the arm is allowed to move, the target has already disappeared and the RNN memory must be relied upon in order for the arm to reach towards the correct target location. The reward in this task is the distance between the positions of end-effector and target, and it tests the ability of DPPO to optimize recurrent network policies. 

\begin{figure}
    \centering
    \includegraphics[width=1.0\textwidth]{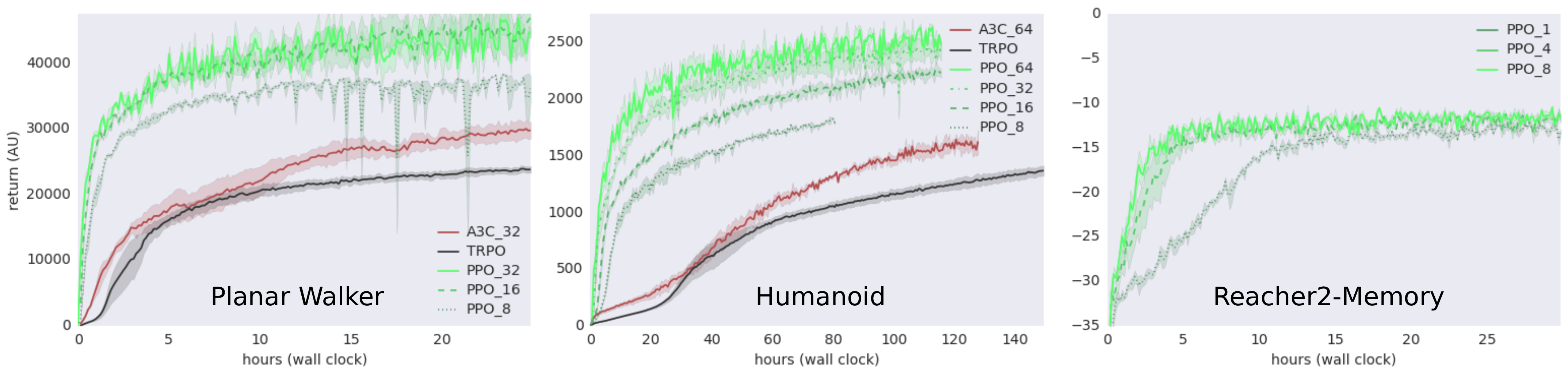}\\
    \caption{\textbf{DPPO benchmark performance} on the \emph{Planar Walker} (\emph{left}), \emph{Humanoid} (\emph{middle}), and \emph{Memory Reacher} (\emph{right}) tasks. In all cases, DPPO achieves performance equivalent to TRPO, and scales well with the number of workers used. The \emph{Memory Reacher}  task demonstrates that it can be used with recurrent networks.}
    \label{fig:AlgorithmComparison}
\end{figure}

\paragraph{Results}
Results depicted in Fig.\ \ref{fig:AlgorithmComparison} show that DPPO achieves performance similar to TRPO and that DPPO scales well with the number of workers used, which can significantly reduce wall clock time. Since it is fully gradient based it can also be used directly with recurrent networks as demonstrated by the \emph{Memory reacher} task. 
DPPO is also faster (in wallclock) than our implementation of A3C when the same number of workers is used.

\section{Methods: environments and models}
\label{sec:Methods}

Our goal is to study whether sophisticated locomotion skills can emerge from simple rewards when learning from varied challenges with a spectrum of difficulty levels.  Having validated our scalable DPPO algorithm on simpler benchmark tasks, we next describe the settings in which we will demonstrate the emergence of more complex behavior.  

\subsection{Training environments}
\label{sec:Parkour}
In order to expose our agents to a diverse set of locomotion challenges we use a physical simulation environment roughly analogous to a platform game, again implemented in Mujoco \cite{todorov2012mujoco}. We procedurally generate a large number of different terrains with a variety of obstacles; a different instance of the terrain and obstacles is generated in each episode.

\paragraph{Bodies} 
We consider three different torque-controlled bodies, described roughly in terms of increasing complexity. \emph{Planar walker}: a simple walking body with 9 DoF and 6 actuated joints constrained to the plane. \emph{Quadruped}: a simple three-dimensional quadrupedal body with 12 DoF and 8 actuated joints. \emph{Humanoid}: a three-dimensional humanoid with 21 actuated dimensions and 28 DoF. The bodies can be seen in action in figures \ref{fig:walkerBehaviors}, \ref{fig:antBehaviors}, and \ref{fig:humanoidBehaviors} respectively.  Note that the \emph{Planar walker} and \emph{Humanoid} bodies overlap with those used in the benchmarking tasks described in the previous section, however the benchmark tasks only consisted of simple locomotion in an open plane.

\paragraph{Rewards}
We keep the reward for all tasks simple and consistent across terrains. The reward consists of a main component proportional to the velocity along the x-axis, encouraging the agent to make forward progress along the track,  plus a small term penalizing torques. For the walker the reward also includes the same box constraints on the pose as in section \ref{sec:distPPO}. For the quadruped and humanoid we penalize deviations from the center of the track, and the humanoid receives an additional reward per time-step for not falling. Details can be found in the supplemental material. We note that differences in the reward functions across bodies are the consequence of us adapting previously proposed reward functions (cf.\ e.g.\ \cite{lillicrap2015continuous,Duan2016Benchmarking}) rather than the result of careful tuning, and while the reward functions vary slightly across bodies we do not change them to elicit different behaviors for a single body.

\paragraph{Terrain and obstacles}
All of our courses are procedurally generated; in every episode a new course is generated based on pre-defined statistics. We consider several different terrain  and obstacle types: (a) \hurdles{}: hurdle-like obstacles of variable height and width that the walker needs to jump or climb over; (b) \gaps{}: gaps in the ground that must be jumped over; (c) \heightfields{}: a terrain with different features such as ramps, gaps, hills, etc.; (d) \walls{}: walls that form obstacles that require walking around, (e) \platforms{}: platforms that hover above the ground which can be jumped on or crouched under. Courses consist of a sequence of random instantiations of the above terrain types within user-specified parameter ranges. 

\begin{figure}
  \centering
  \includegraphics[width=1\textwidth]{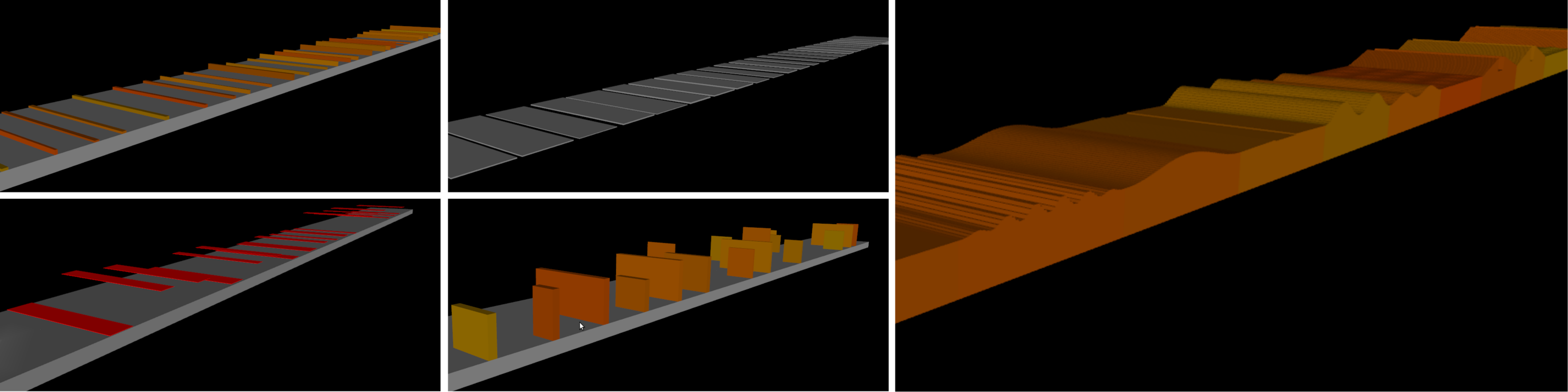}
  \caption{Examples of the terrain types used in the experiments. Left to right and top to bottom: \hurdles{}, \platforms{}, \gaps{}, \walls{}, \heightfields{}.}
  \label{fig:Terrains}
\end{figure}

We train on different types of courses: single-type courses (e.g. gaps only, hurdles only, etc.); mixtures of single-type courses (e.g. every episode a different terrain type is chosen); and mixed terrains (individual courses consisting of more than one terrain type). We consider stationary courses for which the obstacle statistics are effectively fixed over the the length of the course, and ``curriculum'' courses in which the difficulty of the terrain increases gradually over the length of the course. Fig.\ \ref{fig:Terrains} shows a few different course types.

\begin{wrapfigure}{t!}{0.5\textwidth}
  \begin{center}
    \includegraphics[width=0.425\textwidth]{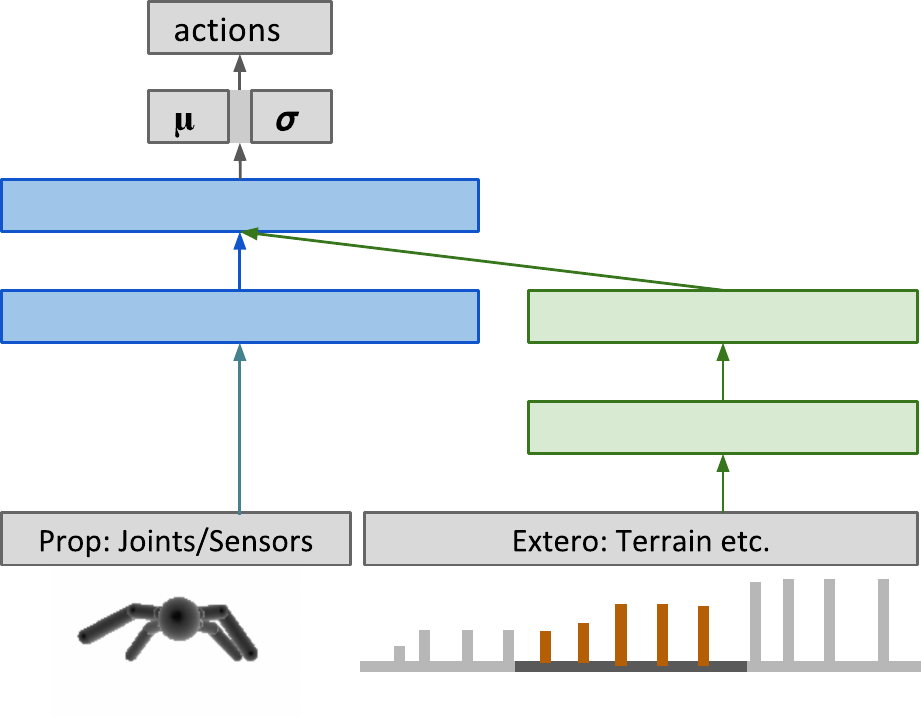}\hspace{0.02\textwidth}
  \end{center}
  \caption{Schematic of the network architecture. We use an architecture similar to \cite{heess2016learning}, consisting of a component processing information local to the controlled body (egocentric information; blue) and a modulatory component that processes environment and task related ``exteroceptive'' information such as the terrain shape (green).}
  \label{fig:Network}
\end{wrapfigure}

\paragraph{Observations} 
The agents receive two sets of observations \cite{heess2016learning}: (1) a set of egocentric, ``proprioceptive'' features containing joint angles and angular velocities; for the \quadruped{} and \humanoid{} these features also contain the readings of a velocimeter, accelerometer, and a gyroscope positioned at the torso providing egocentric velocity and acceleration information, plus contact sensors attached to the feet and legs. The \humanoid{} also has torque sensors in the joints of the lower limbs. (2) a set of ``exteroceptive'' features containing task-relevant information including the position with respect to the center of the track as well as the profile of the terrain ahead. Information about the terrain is provided as an array of height measurements taken at sampling points that translate along the x- and y-axis with the body and the density of which decreases with distance from the body. The \walker{} is confined to the $xz$-plane (i.e. it cannot move side-to-side), which simplifies its perceptual features. See supplemental material for details.

\subsection{Policy parameterization}
\label{sec:Policy}

Similar to \cite{heess2016learning} we aim to achieve a separation of concerns between the basic locomotion skills and terrain perception and navigation. We structure our policy into two subnetworks, 
one of which receives only proprioceptive information, and the other which receives only exteroceptive information. As explained in the previous paragraph with proprioceptive information we refer to information that is independent of any task and local to the body
while exteroceptive information includes a representation of the terrain ahead.
We compared this architecture to a simple fully connected neural network and found that it greatly increased learning speed. Fig.\ \ref{fig:Network} shows a schematic.

\section{Results}
\label{sec:Results}

We apply the Distributed PPO algorithm to a variety of bodies, terrains, and obstacles. Our aim is to establish whether simple reward functions can lead to the emergence of sophisticated locomotion skills when agents are trained in rich environments. We are further interested whether the terrain structure can affect learning success and robustness of the resulting behavior.

\begin{figure}[t!]
  \centering
  \includegraphics[width=1.0\linewidth]{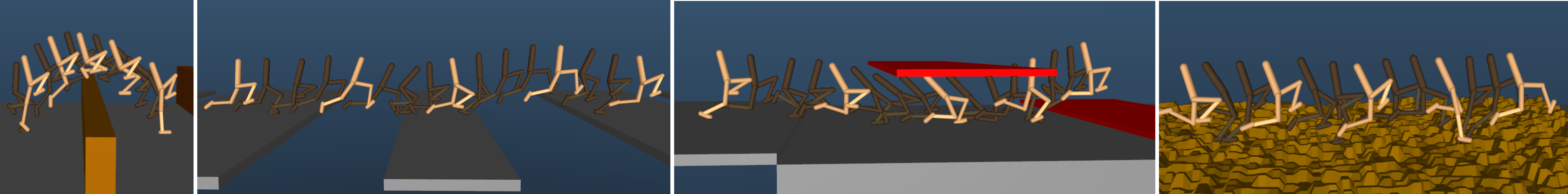}
  \caption{\emph{Walker skills}: Time-lapse images of a representative \walker{} policy traversing rubble; jumping over a hurdle; jumping over gaps and crouching to pass underneath a platform.
  }
  \label{fig:walkerBehaviors}
\end{figure}

\paragraph{Planar Walker}
We train the walker on \hurdles{}, \gaps{}, \platforms{}, and \heightfields{} separately, on a mixed course containing all features interleaved, and on a mixture of terrains (i.e.\ the walker was placed on different terrains in different episodes). It acquired a robust gait, learned to jump over hurdles and gaps, and to walk over or crouch underneath platforms. All of these behaviors emerged spontaneously, without special cased shaping rewards to induce each separate behaviour. Figure \ref{fig:walkerBehaviors} shows motion sequences of the \walker{} traversing a rubble-field, jumping over a hurdle, and over gaps, and crouching under a platform. Examples of the respective behaviors can be found in the supplemental video. The emergence of these skills was robust across seeds. At the end of learning the \walker{} jumped over hurdles nearly as tall as its own body.

\paragraph{Quadruped}

The quadruped is a generally less agile body than the walker but it adds a third dimension to the control problem. We considered three different terrain types: \heightfields{}, \walls{}, \gaps{}{}, and a variation of the \hurdles{} terrain which contained \obstacles{} that can be avoided, and others that require climbing or jumping.

The \quadruped{}, too, learns to navigate most obstacles quite reliably, with only small variations across seeds. It discovers that jumping up or forward (in some cases with surprising accuracy) is a suitable strategy to overcome \hurdles{}, and \gaps{}, and it learns to navigate walls, turning left and right as appropriate -- in both cases despite only receiving reward for moving forward. For the variation of the \hurdles{}-terrain it learns to distinguish between obstacles that it can and / or has to climb over, and those it has to walk around. The \heightfields{} may seem easy but is, in fact, surprisingly hard because the body shape of the \quadruped{}  is poorly suited (i.e. the legs of the quadruped are short compared to the variations in the terrain). Nevertheless it learns strategies to traverse reasonably robustly. Fig.\ \ref{fig:antBehaviors} shows some representative motion sequences; further examples can be found in the supplemental video. 
\begin{figure}[t!]
  \centering
  \includegraphics[width=.55\linewidth]{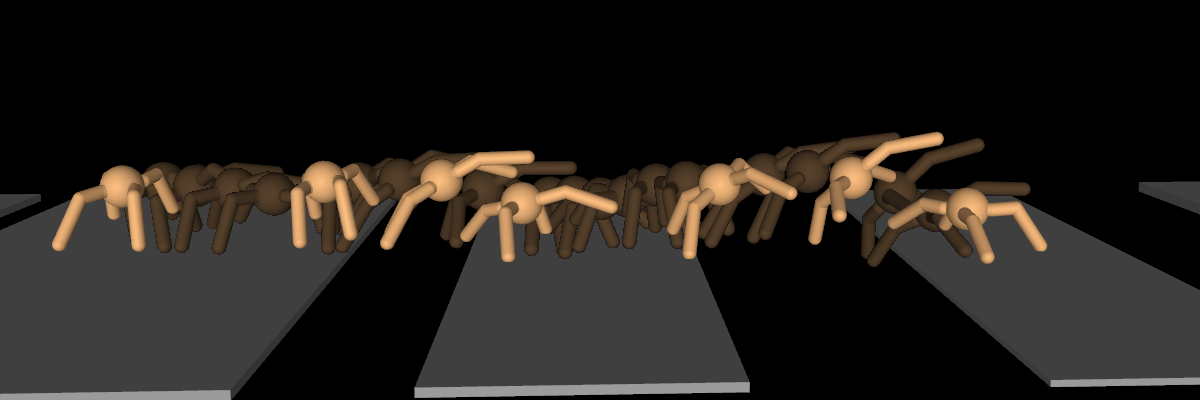}
  \includegraphics[width=.44\linewidth]{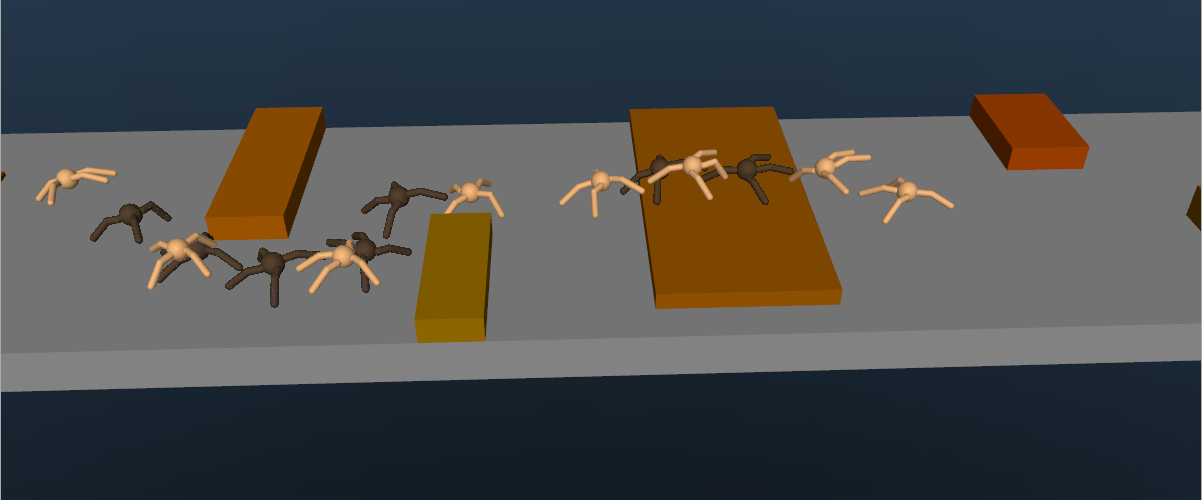}
  \caption{Time-lapse images of a representative \quadruped{} policy traversing gaps (left); and navigating obstacles (right)}
  \label{fig:antBehaviors}
\end{figure}

\begin{figure}[t!]
  \centering
    \includegraphics[width=1.0\linewidth]{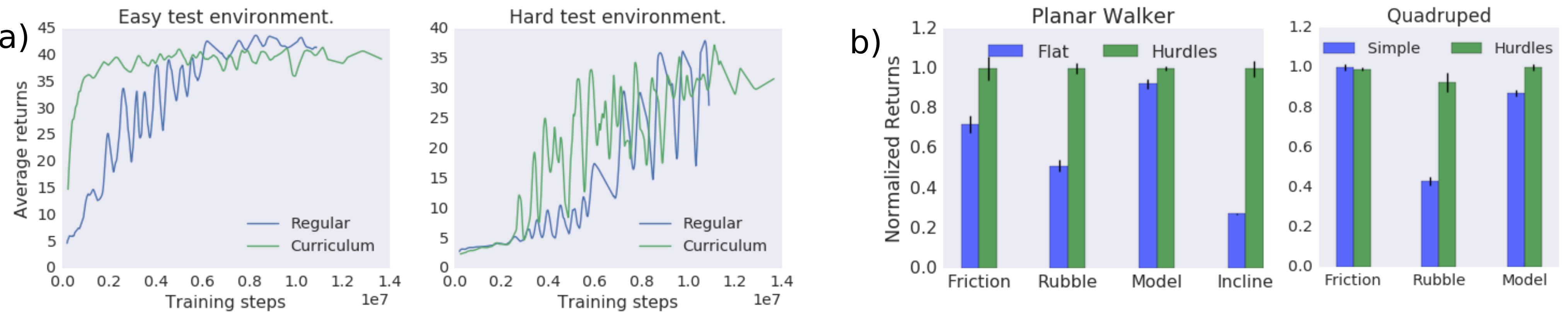}
    \caption{\emph{a) Curriculum training}: Evaluation of policies trained on hurdle courses with different statistics: ``regular'' courses contain arbitrarily interleaved high and low hurdles (blue); ``curriculum'' courses gradually increase hurdle height over the course of the track (green). During training we evaluate both policies on validation courses with low/``easy" hurdles (left) and tall/``hard" hurdles (right). The performance of the policy trained on the curriculum courses increases faster.
    \emph{b) Robustness} of \walker{} policies (left) and \quadruped{} policies (right): We evaluate how training on \hurdles{} (green) increases policy robustness relative to training on flat terrain (blue).
    Policies are assessed on courses with unobserved changes in ground friction, terrain surface (rubble), strength of the body actuators, and incline of the ground plane. There is a notable advantage in some cases for policies trained on the hurdle terrain. All plots show the average returns normalized for each terrain setting.}
  \label{fig:walkerCurriculum}
\end{figure}

\paragraph{Analyses}
We investigate whether the nature of the terrain affects learning. In particular, it is easy to imagine that training on, for instance, very tall hurdles only will not be effective. For training to be successful in our setup it is required that the walker occasionally ``solves'' obstacles by  chance -- and the probability of this happening, is, of course, minuscule when all hurdles are very tall. We verify this by training a \walker{} on two different types of \hurdles{}-terrains. The first possesses stationary statistics with high- and low hurdles being randomly interleaved. In the second terrain the difficulty, as given by the minimum and maximum height of the hurdles, increases gradually over the length of the course. We measure learning progress by evaluating policies during learning on two test terrains, an easy one with shallow hurdles and a difficult one with tall hurdles. Results are shown in Fig.\ \ref{fig:walkerCurriculum}a for a representative \walker{} policy. The policy trained on the terrain with gradually increasing difficulty improves faster than the one trained on a stationary terrain. 

We further study whether training on varying terrains leads to more robust gaits compared to usual task of moving forward on a plane. To this end we train \walker{}  and \quadruped{} policies on a flat course as well as on the (more challenging) hurdles. We then evaluate representative policies from each experiment with respect to their robustness to (a) unobserved variations in surface friction, (b) unobserved rumble-strips, (c) changes in the model of the body, (d) unobserved inclines / declines of the ground. Results depicted in Fig.\ \ref{fig:walkerCurriculum}b show a trend of training on hurdles increasing robustness on other forms of unobserved variation in the terrain.

\paragraph{Humanoid} Our final set of experiments considers the 28-DoF \humanoid{}, a considerably more complex body than \walker{} and \quadruped{}. The set of terrains is qualitatively similar to the ones used for the other bodies, including \gaps{}, \hurdles{}, a \heightfields{}, as well as the \walls{}. We also trained agents on mixtures of the above terrains.

As for the previous experiments we considered a simple reward function, primarily proportional to the velocity along the x-axis (see above).
We experimented with two alternative termination conditions: (a) episodes were terminated when the minimum distance between head and feet fell below 0.9m; (b) episodes were terminated when the minimum distance between head and ground fell below 1.1m.

In general, the humanoid presents a considerably harder learning problem largely because with its relatively large number of degrees of freedoms it is prone to exploit redundancies in the task specification and / or to get stuck in local optima, resulting in entertaining but visually unsatisfactory gaits. Learning results tend to be sensitive to the particular algorithm, exploration strategy, reward function, termination condition, and weight initialization. 

The results we obtained for the humanoid were indeed much more diverse than for the other two bodies, with significant variations across seeds for the same setting of the hyperparameters. Some of the variations in the behaviors were associated with differences in learning speed and asymptotic performance (suggesting a local optimum); others were not (suggesting alternative solution strategies).

\begin{figure}[t!]
  \centering
  \includegraphics[width=1.0\linewidth]{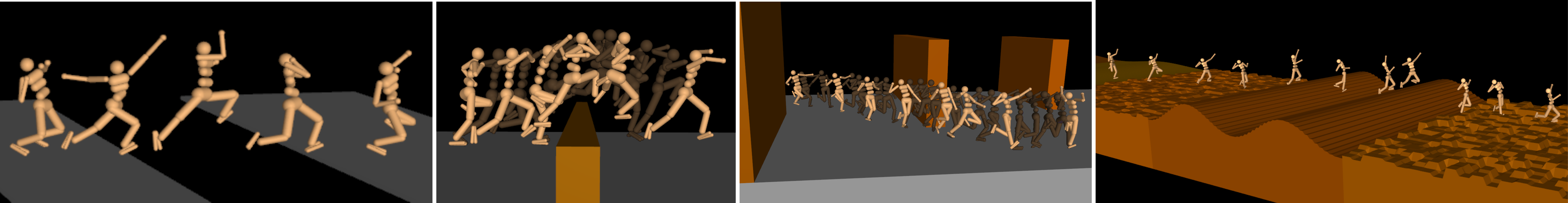}
\caption{Time lapse sequences of the \humanoid{} navigating different terrains}
  \label{fig:humanoidBehaviors}
\end{figure}

Nevertheless we obtained for each terrain several well performing agents, both in terms of performance and in terms of visually pleasing gaits. Fig.\ \ref{fig:humanoidBehaviors} shows several examples of agents trained on \gaps{}, \hurdles{}, \walls{}, and \heightfields{}. As in the previous experiments the terrain diversity and the inherent curriculum led the agents to discover robust gaits, the ability to overcome obstacles, to jump across gaps, and to navigate slalom courses.  We highlight several solution strategies for each terrain in the supplemental video, including less visually appealing ones.  To test the robustness of the learned behaviors we further constructed two test courses with (a) statistics rather different from the training terrains and (b) unobserved perturbations in the form of see-saws and random forces applied to the \humanoid{}'s torso, which is also presented in the video. Qualitatively we see moderately large levels of robustness to these probe challenges (see supplemental video).

\section{Related work}

Physics-based character animation is a long-standing and active field that has produced a large body of work with impressive results endowing simulated characters with locomotion and other movement skills (see \cite{geijtenbeek2012interactive} for a review). For instance, \cite{TerrainRunner} show sophisticated skill sequencing for maneuvering obstacles on a parametric terrain, while \cite{wu2010terrain,mordatch2010robust,mordatch2012discovery} demonstrate how terrain adaptive behaviors or other skilled movements can emerge as the result of optimization problems. While there are very diverse approaches, essentially all rely on significant prior knowledge of the problem domain and many on demonstrations such as motion capture data. 

Basic locomotion behaviors learned end-to-end via RL have been demonstrated, for instance, by \cite{Schulman2015Trust,lillicrap2015continuous,schulman2015high,heess2015learning} or guided policy search \cite{levine2014learning}. Locomotion in the context of higher-level tasks has been considered in \cite{heess2016learning}. Terrain-adaptive locomotion with RL has been demonstrated by \cite{terrain_adaptive}, but they still impose considerable structure on their solution. Impressive results were recently achieved with learned locomotion controllers for a 3D humanoid body \cite{2017-TOG-deepLoco}, but these rely on a domain-specific structure and human motion capture data to bootstrap the movement skills for navigating flat terrains.

The idea of curricula is long-standing in the machine learning literature (e.g.\ \cite{bengio2009curriculum}). It has been exploited for learning movement skills for instance  by \cite{karpathy2012curriculum}.  The present work combines and develops elements from many of these research threads, but pushes uniquely far in a particular direction -- using simple RL rewards and curriculum training to produce adaptive locomotion in challenging environments while imposing only limited structure on the policy and behavior.

\section{Discussion}

We have investigated the question whether and to what extent training agents in a rich environment can lead to the emergence of behaviors that are not directly incentivized via the reward function. This departs from the common setup in control where a reward function is carefully tuned to achieve particular solutions. Instead, we use deliberately simple and generic reward functions but train the agent over a wide range of environmental conditions.  Our experiments suggest that training on diverse terrain can indeed lead to the development of non-trivial locomotion skills such as jumping, crouching, and turning for which designing a sensible reward is not easy. While we do not claim that environmental variations will be sufficient, we believe that training agents in richer environments and on a broader spectrum of tasks than is commonly done today is likely to improve the quality and robustness of the learned behaviors -- and also the ease with which they can be learned. In that sense, choosing a seemingly more complex environment may actually make learning easier.

\subsubsection*{Acknowledgments}

We thank Joseph Modayil and many other colleagues at DeepMind for helpful discussions and comments on the manuscript.

\newpage

\setlength{\bibsep}{1pt plus 0.3ex}
{\small 

}
\newpage
\appendix

\title{Supplmentary material}



\maketitle
 
\section{Distributed PPO}

\subsection{Algorithm details}
Pseudocode for the Distributed PPO algorithm is provided in Algorithm Boxes \ref{alg:DPPO_chief} and \ref{alg:DPPO_workers}. $W$ is the number of workers; $D$ sets a threshold for the number of workers whose gradients must be available to update the parameters. $M, B$ is the number of sub-iterations with policy and baseline updates given a batch of datapoints. $T$ is the number of data points collected per worker before parameter updates are computed. $K$ is the number of time steps for computing $K$-step returns and truncated backprop through time (for RNNs)

\begin{algorithm}
\caption{Distributed Proximal Policy Optimization (chief)}
\label{alg:DPPO_chief}
\footnotesize
\begin{algorithmic}
    \FOR {$i \in \{1, \cdots, N\}$}
        \FOR {$j \in \{1, \cdots, M\}$ }
            \STATE Wait until at least $W-D$ gradients wrt. $\theta$ are available
            \STATE average gradients and update global $\theta$
        \ENDFOR
        \FOR { $j \in \{1, \cdots, B\}$ }
            \STATE Wait until at least $W-D$ gradients wrt.\ $\phi$ are available
            \STATE average gradients and update global $\phi$
        \ENDFOR
	\ENDFOR
\end{algorithmic}
\end{algorithm}

\begin{algorithm}
\caption{Distributed Proximal Policy Optimization (worker)}
\label{alg:DPPO_workers}
\footnotesize
\begin{algorithmic}
    \FOR {$i \in \{1, \cdots, N\}$}
        \FOR {$w \in \{1, \cdots T/K\}$}
            \STATE Run policy $\pi_{\theta}$ for $K$ timesteps, collecting $\{s_t,a_t,r_t\}$ for $t \in \{ (i-1)K, \dots, iK-1\}$
            \STATE Estimate return $\hat{R}_t = \sum_{t= (i-1)K}^{iK-1} \gamma^{t-(i-1)K} r_{t} + \gamma^K V_\phi(s_{iK})$
            \STATE Estimate advantages $\hat{A}_t = \hat{R}_t - V_\phi(s_t)$
            \STATE Store partial trajectory information
        \ENDFOR
        \STATE $\pi_\mathrm{old} \leftarrow \pi_\theta$
        \FOR {$m \in \{1, \cdots, M\}$}
            \STATE $J_{PPO}(\theta) =  \sum_{t=1}^T \frac{\pi_{\theta}(a_t|s_t)}{\pi_{old}(a_t|s_t)} \hat{A}_t - \lambda \mathrm{KL}[\pi_{old}|\pi_{\theta}] - 
            \xi \mathrm{max}(0, \mathrm{KL}[\pi_{old}|\pi_{\theta}] - 2\mathrm{KL}_{target})^2$
            \IF {$\mathrm{KL}[\pi_{old}|\pi_{\theta} > 4\mathrm{KL}_{target} $}
            \STATE break and continue with next outer iteration $i+1$
            \ENDIF
            \STATE Compute $\nabla_\theta J_{PPO}$ 
            \STATE send gradient wrt.\ to $\theta$ to chief
            \STATE wait until gradient accepted or dropped; update parameters
            
		\ENDFOR
		\FOR {$b \in \{1, \cdots, B\}$}
            \STATE $L_{BL}(\phi) =  -\sum_{t=1}^T (\hat{R}_t - V_\phi(s_t))^2$
            \STATE Compute $\nabla_\phi L_{BL}$ 
            \STATE send gradient wrt.\ to $\phi$ to chief
            \STATE wait until gradient accepted or dropped; update parameters
		\ENDFOR
		\IF {$\mathrm{KL}[\pi_{old}|\pi_{\theta}] > \beta_{high}\mathrm{KL}_{target}$}
		\STATE  $\lambda \leftarrow \tilde{\alpha} \lambda$
		\ELSIF {$\mathrm{KL}[\pi_{old}|\pi_{\theta}] < \beta_{low}\mathrm{KL}_{target}$}
		\STATE  $\lambda \leftarrow  \lambda/\tilde{\alpha} $
		\ENDIF
	\ENDFOR	
\end{algorithmic}
\end{algorithm}

\paragraph{Normalization} Following \cite{schulmanPPOcode} we perform the following normalization steps: 
\begin{enumerate}
\item We normalize observations (or states $s_t$) by subtracting the mean and dividing by the standard deviation using the statistics aggregated over the course of the entire experiment.
\item We scale the reward by a running estimate of its standard deviation, again aggregated over the course of the entire experiment.
\item We use per-batch normalization of the advantages. 
\end{enumerate}

\paragraph{Sharing of algorithm parameters across workers}
In the distributed setting we have found it to be important to share relevant statistics for data normalization across workers. Normalization is applied during data collection and statistics are updated locally after every environment step. Local changes to the statistics are applied to the global statistics after data collection when an iteration is complete (not shown in pseudo-code). 
The time-varying regularization parameter $\lambda$ is also shared across workers but updates are determined based on local statistics based on the average KL computed locally for each worker, and applied separately by each worker with an adjusted $\tilde{\alpha} = 1 + (\alpha - 1) / K$.

\paragraph{Additional trust region constraint}
We also adopt an additional penalty term that becomes active when the KL exceeds the desired change by a certain margin (the threshold is $2\mathrm{KL}_{target}$ in our case). In our distributed implementation this criterion is tested and applied on a per-worker basis.

Stability is further improved by early stopping when changes lead to too large a change in the KL. 

\subsection{Algorithm comparison}

TRPO has been established as a robust algorithm that learns high-performing policies and requires little parameter tuning. Our primary concern was therefore whether DPPO can achieve results comparable to TRPO. Secondarily, we were interested in whether the algorithm scales to large numbers of workers and allows speeding up experiments where large numbers of data points are required to obtain reliable gradient estimates.
We therefore compare to TRPO in a regime where a large number number samples is used to compute parameter updates ($N=100000$). For simple tasks we expect TRPO to produce good results in this regime (for the benchmark tasks a smaller $N$ would likely be sufficient).

For DPPO we perform a coarse search over learning rate for policy and baseline.  All experiments in section \ref{sec:EvalDPPO} use the same learning rates ($0.00005$ and $0.0001$ respectively.) In each iteration we use batches of size of 64000 (walker),  128000 (humanoid), and 24000 (reacher) time steps. Data collection and gradient computation are distributed across varying numbers of workers. Due to early termination this number is sometimes smaller (when an episode terminates early the remaining steps in the current unroll window of length $K$ are being ignored during gradient calculation). An alternative point of comparison would be to use a fixed overall number of time steps and vary the number of time steps per worker.

Networks use $\tanh$ nonlinearities and parameterize the mean and standard deviation of a conditional Gaussian distribution over actions. Network sizes were as follows: \walker{}: 300,200; \humanoid: 300,200,100; \reacher: 200; and 100 \emph{LSTM} units.

For A3C with continuous actions we also perform a coarse search over relevant hyper parameters, especially the learning rate and entropy cost. Due to differences in the code base network architectures were not exactly identical to those used for DPPO but used the same numbers of hidden units.

We note that a like-for-like comparison of the algorithms is difficult since they are implemented in different code bases and especially for distributed algorithms performance in wall clock time is affected both by conceptual changes to the algorithm as well as by implementation choices. A more careful benchmarking of several recent high-throughput algorithms will be the subject of future work.

\section{Additional experimental details}
\subsection{Observations}

For all courses terrain height (and platform height where applicable) was provided as a heightfield where each "pixel" indicates the height of the terrain (platform) within a small region. This heightfield was then sampled at particular points relative to the position of the agent.

\paragraph{Planar walker}
The exteroceptive features for the planar walker consist of sampling points of the terrain and, where applicable, platform height. There were $50$ equally spaced points along the $x$-axis starting $2m$ behind the agent and extending $8m$ ahead. Platform height was represented separately from terrain height with a separate set of sampling points. In addition the exteroceptive features contained the height of the walker body above the ground (measured at its current location) as well as the difference between the agents position and the next sampling grid center (the intention behind this last input is to resolve the aliasing arising from the piece-wise constant terrain representation with finite sampling).

\paragraph{Quadruped \& Humanoid}
The \quadruped{} and \humanoid{} use the same set of  exteroceptive features, effectively a two-dimensional version of what is used for the walker. The sampling points are placed on a variable-resolution grid and range from $1.2m$ behind the agent to $5.6m$ ahead of it along the $x$-axis as well as $4m$ to the left and to the right. To reduce dimensionality of the input data sampling density decreases with increasing distance from the position of the body. In addition to the height samples the exteroceptive features include the height of the body above the ground, and the x and y distance of the walker body to the next sampling grid center (to reduce aliasing; see above).

\subsection{Rewards}

\paragraph{Planar walker}
$r = 10.0v_x +0.5n_z - \left |\Delta_h-1.2\right | -10.0I[\Delta_h < 0.3] - 0.1 \|u\|^2$

Here $n_z$ is the projection of the $z$-axis of the torso coordinate frame onto the $z$-axis of the global coordinate frame (this value varies from 1.0 to -1.0) depending on whether the \walker{}'s torso is upright or upside down. $\Delta_h$ is the height of the \walker{}'s torso above the feet. $I[\cdot]$ is the indicator function. $v_x$ is the velocity along the $x$-axis.

\paragraph{Quadruped}
$r  = v_x + 0.05 n_z - 0.01\|u\|^2$ where $n_z$ is the projection of the $z$-axis of the torso coordinate frame onto the $z$-axis of the global coordinate frame (this value varies from 1.0 to -1.0) depending on whether the \quadruped{} is upright or upside down. 

\paragraph{Humanoid}
$r= \min(v_x,v_{\mathrm{max}}) - 0.005(v_x^2 + v_y^2) -0.05y^2 - 0.02\|u\|^2 + 0.02$ where $v_{\mathrm{max}}$ is a cutoff for the velocity reward which we usually set to $4m/s$.

\end{document}